\documentclass[11pt]{article}

\usepackage[final]{acl}

\usepackage{times}
\usepackage{latexsym}

\usepackage[T1]{fontenc}

\usepackage[utf8]{inputenc}

\usepackage{microtype}

\usepackage{inconsolata}

\usepackage{graphicx}

\usepackage{float}
\usepackage{pifont}
\usepackage{subfigure}
\usepackage{multirow}
\usepackage{threeparttable}
\usepackage[ruled,linesnumbered]{algorithm2e}
\usepackage{scalerel}
\usepackage{booktabs}
\usepackage{amsfonts}
\usepackage{amsmath}

\def\modelname{PPL}

\newcommand{\squishlist}{
   \begin{list}{$\bullet$}
    { \setlength{\itemsep}{0pt}      \setlength{\parsep}{2pt}
      \setlength{\topsep}{1pt}       \setlength{\partopsep}{0pt}
      \setlength{\leftmargin}{1em} \setlength{\labelwidth}{1em}
      \setlength{\labelsep}{0.5em} } }

\newcommand{\squishend}{
    \end{list}  }

\title{Learning Personalized Prompts for Healthcare Guidance}

\author{
 \textbf{Ruize Shi\textsuperscript{1*}},
 \textbf{Hong Huang\textsuperscript{1*$\dagger$}},
 \textbf{Wei Zhou\textsuperscript{1*}},
 \textbf{Kehan Yin\textsuperscript{1*}},
 \textbf{Kai Zhao\textsuperscript{1,2}},
 \textbf{Yun Zhao\textsuperscript{1,2,3}}
\\
\textsuperscript{1}Huazhong University of Science and Technology \\
\textsuperscript{2}Tongji Medical College\\
\textsuperscript{3}Maternal and Child Health Hospital of Hubei Province\\
\texttt{\{rzshi, honghuang, weizhou2021, kehanyin, kai$\_$zhao, 2019fy0004\}@hust.edu.cn}
}

\begin{document}
\maketitle

\renewcommand{\thefootnote}{*}
\footnotetext[1]{Authors are affiliated with the National Engineering Research Center for Big Data Technology and System, Services Computing Technology and System Lab, Cluster and Grid Computing Lab, School of Computer Science and Technology, Huazhong University of Science and Technology.}
\renewcommand{\thefootnote}{$\dagger$}
\footnotetext[2]{Hong Huang is the corresponding author.}
\renewcommand{\thefootnote}{\arabic{footnote}}

\begin{abstract}
The rapid development of {\em large language models} (LLMs) has transformed many industries, including healthcare. In practice, hospitals and patients increasingly seek LLM-based systems capable of interpreting personal health records and providing healthcare guidance. However, existing approaches mainly rely on general medical knowledge and often fail to account for individual variability, limiting their ability to provide personalized guidance.
To address this, we propose {\em personalized prompt learning} ({\modelname}), a framework that learns individualized prompts to guide LLMs in generating personalized healthcare recommendations. {\modelname} constructs initial personalized prompts by leveraging both self-informed patient information and peer-informed signals derived from clinically similar cases. These prompts are then refined using {\em reinforcement learning} (RL) to better align the generated responses with physician recommendations written for each patient. {\modelname} operates with hard prompts, enabling seamless integration with proprietary LLMs without modifying the underlying models. We evaluate {\modelname} on real-world obstetrics and gynecology data. The results show that our approach produces more personalized healthcare guidance and wins 97 out of 100 comparisons in expert evaluation, demonstrating its potential for broader healthcare applications. Our code is publicly available at \url{https://github.com/CGCL-codes/PPL}.
\end{abstract}

\section{Introduction}
In recent years, the explosive growth of generative artificial intelligence has profoundly impacted various industries, drawing widespread attention to {\em large language models} (LLMs) \cite{llmecommerce,toolllmsurvey,llmsurvey}. LLMs have been applied in many fields, including healthcare, where they provide medical knowledge and valuable insights \cite{llmmedicineprospect,llmmedical}. Increasingly, patients and healthcare providers rely on LLM-based systems to interpret personal health records and obtain medical guidance \cite{llmmedicinesurvey1,llmmedicinesurvey2}.

\begin{figure}[t]
    \centering
    \includegraphics[width=0.99\linewidth]{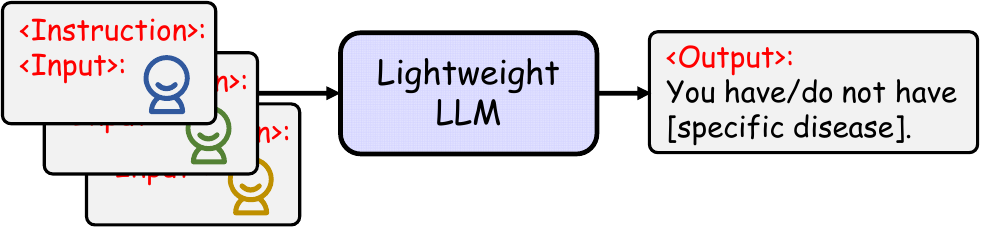}
    \caption{Prior works on personalized medical LLMs often target single disease settings, where personalization mainly comes from differences in patient inputs while overlooking personalization at the prompt level.}
    \label{fig:previouswork}
\end{figure}

This growing adoption highlights an urgent demand for personalization in healthcare guidance with LLMs, as both patients and healthcare providers require guidance tailored to individual needs \cite{malp,pllmsurvey2}. While some studies claim to have developed such LLMs \cite{personalizeddiabetes,personalizedprostatecancer,malp}, these models are typically restricted to a single disease and personalize only based on limited variations in patient data, as shown in Figure \ref{fig:previouswork}. Specifically, they focus on fine-tuning \cite{prefixtuning,prompttuning} lightweight LLMs \cite{llama,chatglm} with knowledge about a specific disease, without the ability to adapt to individual patients. Thus, prior approaches provide only limited forms of personalization and cannot fully capture patient level variability in real-world healthcare scenarios. 

As such, there remains a pressing need for effective mechanisms to enable personalization in LLM healthcare guidance, raising two key challenges: {\em (1) How to support multiple diseases while adapting to individual patients?} Most existing works rely on models designed for a single disease and provide only limited personalization based on variations in patient inputs. In reality, patients present with a wide range of conditions. Thus, enabling LLMs to provide guidance across diverse diseases while still tailoring responses to individual patients remains challenging. {\em (2) How can personalization be achieved automatically?} Current methods typically extract limited patient information and insert it into fixed prompt templates to guide LLM outputs. Such approaches cannot easily adapt to evolving patient records, contextual differences, or variations across underlying LLMs. Achieving fine-grained personalization therefore requires methods that can automatically refine and adjust the prompts used to guide the LLM for each patient.

To this end, we propose {\em personalized prompt learning} ({\modelname}), a framework that learns personalized prompts to guide LLMs in generating tailored healthcare guidance. {\modelname} analyzes patient history to extract personalized information and incorporates insights from similar patients via collaborative filtering. Using self-informed and peer-informed insights, it constructs coarse-grained personalized prompts, which are then refined through {\em reinforcement learning} (RL), where physician replies for each patient serve as the optimization signal. The final prompts are fed into a high-quality LLM to guide the generation of customized responses. Note that prompt refinement considers the characteristics of the downstream LLM, and the personalized prompts are hard prompts \cite{hardprompt1,hardprompt2}. Thus, {\modelname} can be reused across different LLMs and seamlessly integrated with proprietary LLMs with extensive medical knowledge. Our contributions are summarized as follows:

\squishlist
    \item We study how personalized prompts can improve healthcare guidance generated by LLMs. While previous methods achieve only basic personalization, our method enables more refined guidance customized to individual patients.
    \item We propose a framework named {\modelname}, which uses RL to automatically adjust prompts based on patient information and downstream LLMs, guiding proprietary LLMs with extensive medical knowledge to offer tailored responses.
    \item Experiments on real-world obstetrics and gynecology data show that {\modelname} significantly enhances personalization in proprietary LLMs and outperforms fine-tuned LLMs, as confirmed by both automatic metrics and expert evaluation.
\squishend

\section{Related Work}

\subsection{Personalized LLMs}

Recently, personalization in LLMs has attracted attention \cite{pllmsurvey1,pllmsurvey2}, particularly in recommendation systems \cite{llmrecommendationsurvey}. Some researchers focus on combining user history to improve recommendation quality \cite{pllm,prs,pmg}, while others leverage reinforcement learning from human feedback to understand users' needs and enhance personalization \cite{prlhf,psoups}. However, these methods often rely on item embedding, making them difficult to apply in healthcare. Meanwhile, many LLMs have been applied in the healthcare domain and have shown promising results \cite{personalizedhealth,llmmedicinesurvey1,llmmedicinesurvey2}. Some studies attempt to develop personalized medical LLMs, yet most focus on advice or treatment for specific diseases \cite{personalizedoncology,personalizeddiabetes,personalizedplan}. Specifically, they fine-tune LLMs to learn general treatment strategies for a specific disease, achieving personalization at the disease level rather than tailoring prompts to individual patients. Moreover, they usually fine-tune lightweight LLMs, making limited use of proprietary LLMs with more extensive medical knowledge.

\begin{figure*}[ht]
    \centering
    \includegraphics[width=0.99\linewidth]{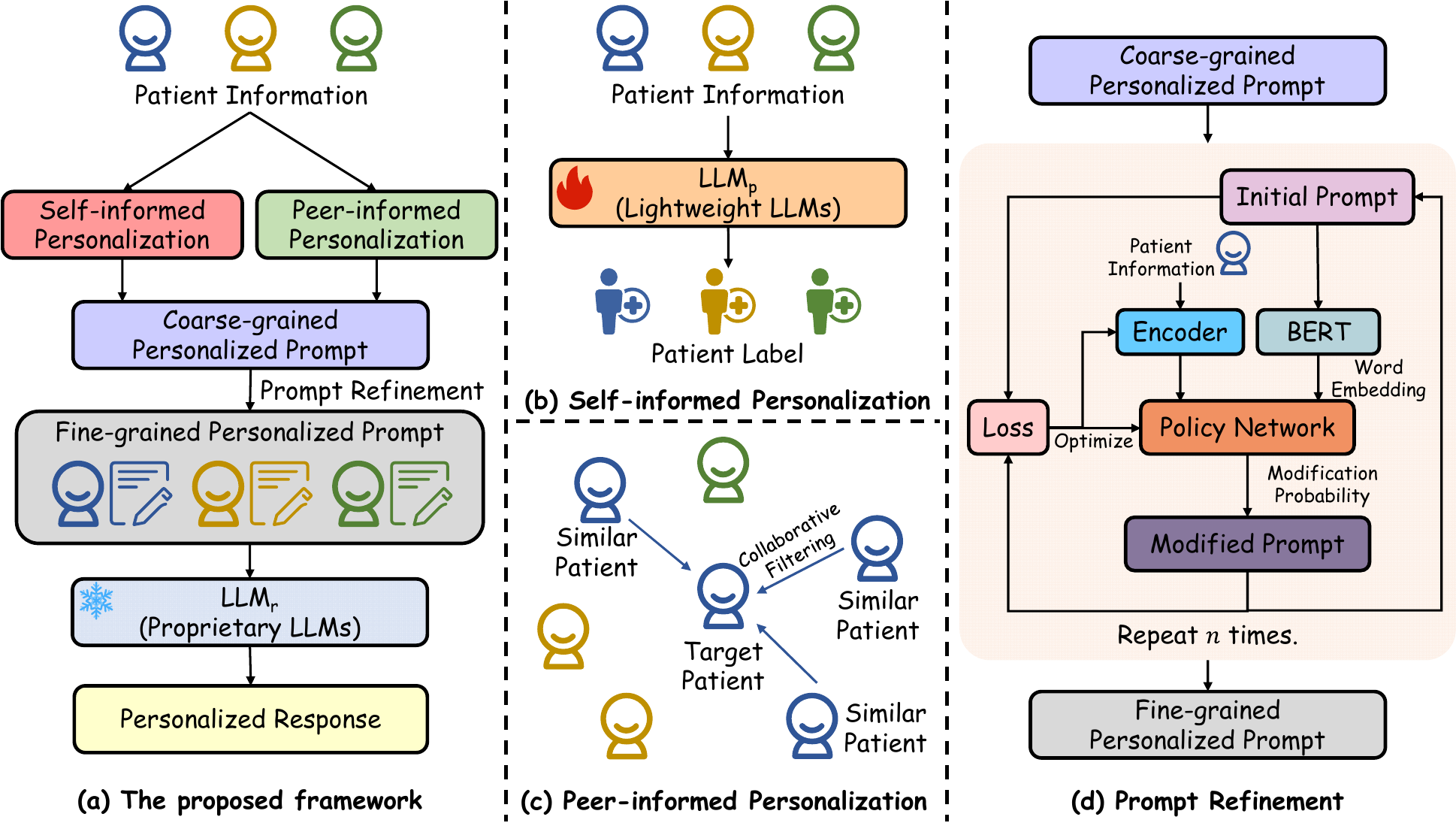}
    \caption{Overview of {\modelname}. The framework constructs personalized prompts using patient records and similar cases, and refines them through RL to guide LLMs in generating personalized responses.}
    \label{fig:model}
\end{figure*}

\subsection{Hard Prompt Optimization}
Early work \cite{autoprompt} on hard prompt optimization is conducted on pre-trained models. Later, GrIPS \cite{grips} and FluentPrompt \cite{fluentprompt} are applied to GPT-2 \cite{gpt2}, where GrIPS is a gradient-free approach operating at phrase level, while FluentPrompt uses Langevin dynamics. PEZ \cite{pez} optimizes hard prompt through soft prompt as intermediaries. Several works explore modifying hard prompts through RL \cite{rlsurvey1,rlsurvey2}, and RLPrompt \cite{rlprompt} is among the earliest efforts. Based on RLPrompt, PIN \cite{pin} enhances prompt interpretability by sparse Tsallis entropy regularization \cite{ster}. PRewrite \cite{prewrite} further uses a retrieval strategy to determine the most effective prompt. Rewriter \cite{rewriter} combines supervised learning with RL to rewrite prompts. Recent work \cite{fluentprompt*} explores optimizing prompts in continuous representation spaces. However, these methods focus on optimal prompts for specific tasks, rather than designing personalized prompts tailored to individual users.

\section{Method Preparation}

We denote a healthcare dataset as $\mathcal{H} = \{\mathcal{X}, \mathcal{Y}, \mathcal{R}\}$, where $\mathcal{X} = (x_{1}, x_{2}, \cdots, x_N)$ contains $N$ patients, $\mathcal{Y} = (y_{1}, y_{2}, \cdots, y_N)$ denotes patient labels, such as diseases, or a normal state, and $\mathcal{R} = (r_{1}, r_{2}, \cdots, r_N)$ is physician recommendations written for each patient, which serve as reference responses and reflect personalized healthcare guidance. Each patient $x_{i} \in \mathbb{R}^{N_{x_{i}} \times N_{m}}$, where $N_{x_{i}}$ is the health check counts and $N_{m}$ denotes the health examination metrics. Since patients may have different numbers of health checks, we use subscripts for differentiation.

Based on this, we extract patient specific information from health records and leverage signals from similar patients. RL is then used to refine the prompt so that the generated responses better align with physician recommendations, thereby achieving patient level personalization.

Personalized prompt is $P = (p_{1}, p_{2}, \cdots, p_{N_{p}})$, where $p_i$ represents a word. The prompt is represented in textual form, allowing it to be compatible with closed-source LLMs without modifications to their architectures. Furthermore, $P_{i} \neq P_{j}, i \neq j$ for different patients, meaning that each patient is associated with a distinct prompt. In addition, the prompt length may vary across patients.

\section{The Proposed Method}

We provide a detailed introduction to our proposed {\modelname}. As illustrated in Figure \ref{fig:model}, {\modelname} guides LLMs to generate personalized responses via tailored prompts. 
{\modelname} first employs a predictor to extract personalized information from the patient medical records (self-informed personalization). It then retrieves information from similar patients to further enrich the personalization (peer-informed personalization). Using both sources of information, {\modelname} generates an initial coarse-grained personalized prompt. Through RL, these prompts are iteratively refined into fine-grained prompts tailored to each individual, where the reward encourages the generated responses to align with tailored responses written by physicians for each patient. The proposed {\modelname} does not modify the parameters or architecture of the underlying LLM, enabling seamless integration with proprietary LLMs that possess extensive medical knowledge.

\subsection{Extraction of Personalized Information}

{\modelname} leverages personalized information to build tailored prompts that guide LLMs in generating individualized responses. To achieve this, we extract personalized information from two complementary sources: self-informed personalization derived from the target patient's historical health records, and peer-informed personalization obtained from clinically similar patients.

\begin{figure}[t]
    \centering
    \includegraphics[width=0.99\linewidth]{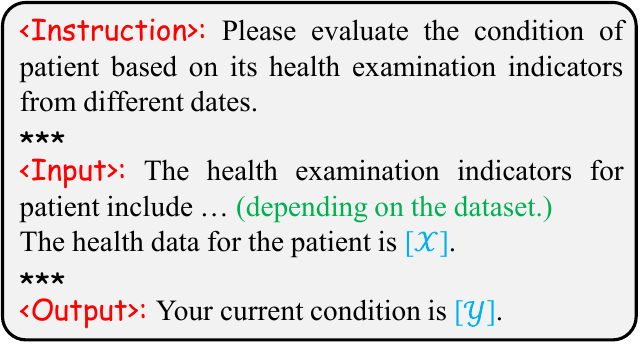}
    \caption{Example prompt for the predictor.}
    \label{fig:predictorprompt}
\end{figure}

{\noindent\bf Self-informed Personalization.} In our work, a patient $x_{i} \in \mathbb{R}^{N_{x_{i}} \times N_{m}}$ consists of multiple health check results from different dates, and the number of checks varies for each individual. To efficiently handle this variable-length data, we use LoRA \cite{lora} to fine-tune LLMs as predictors, denoted as $LLM_p$. Specifically, given a patient's health check data and label, we format the data as shown in Figure \ref{fig:predictorprompt} for instruction tuning. 

After fine-tuning, $LLM_p$ is able to make predictions based on the historical health check data of patient $\mathcal{X}$, which can be formally defined as:
\begin{equation}\label{eq_predict}
    \hat{\mathcal{Y}} = LLM_{p}(\mathcal{X}),
\end{equation}
where $\widehat{\mathcal{Y}}$ denotes the predicted label. 
Despite varying numbers of health checkups, the historical health data can be treated as time series. We evaluate several predictors, including traditional time series methods and LLMs (see Appendix \ref{predictorstudy}), and ultimately adopt LLaVA1.5-7B \cite{llava1.5} as our predictor, since it delivers the best results even though it is primarily a multimodal LLM.

{\noindent\bf Peer-informed Personalization.} To further enrich personalized information, in addition to directly using the original patient data for predictions, {\modelname} extracts information from similar patients. First, {\modelname} utilizes an encoder to map the patient data into the same space. Formally, for each patient $x_i$, it can be represented as follows:
\begin{align}
    \hat{x_{i}} & = encoder(x'_{i}), \label{eq_encoder}\\
    x'_{i} & = padding(\sum^{N_{x_{i}}}_{j=1} \|x_{ij}), \label{eq_padding}
\end{align}
where $padding(\cdot)$ refers to filling in missing values, as the number of health checkups $N_{x_{i}}$ varies between patients, and we apply zero padding here for simplicity. $\hat{x_i} \in \mathbb{R}^{1 \times d}$ denotes the mapped data, and $d$ indicates the mapped space dimension. Furthermore, $x_{ij}$ represents the $j$-th health check result of patient $x_i$, and $\|$ signifies the concatenation. The primary purpose of the encoder is to standardize the dimensions of the input data. To ensure efficiency, we employ the {\em multi-layer perceptron} (MLP) \cite{mlp} here.

{\modelname} then leverages the encoded data to retrieve similar patients to the target patient based on collaborative filtering \cite{collaborativefiltering}. In detail, since the labels in the training set are known, we calculate the cosine similarity between the target patient $x_i$ and each patient $x_j$ in the training set, which is computed as follows:
\begin{equation}\label{eq_sim}
    w_{ij} = \frac{\hat{x_i} \cdot \hat{x_j}}{\| \hat{x_i} \|_{2} \cdot \| \hat{x_j} \|_{2}},
\end{equation}
where $w_{ij}$ indicates the similarity between patients $x_i$ and $x_j$, and $\| \cdot \|_{2}$ is the Euclidean norm. {\modelname} leverages the top-$k$ similar patients to enhance the target patient's personalized information for constructing a tailored prompt. $k$ is a hyper-parameter, which we explore in detail in Section \ref{hyperparameterstudy}.

\subsection{Generation of Personalized Prompt}

Based on the self-informed and peer-informed personalized information, {\modelname} constructs tailored prompts to guide LLMs in generating customized responses. We first format and create an initial prompt, then refine this prompt through RL, ultimately producing unique prompts for each patient.

\begin{figure}[t]
    \centering
    \includegraphics[width=0.99\linewidth]{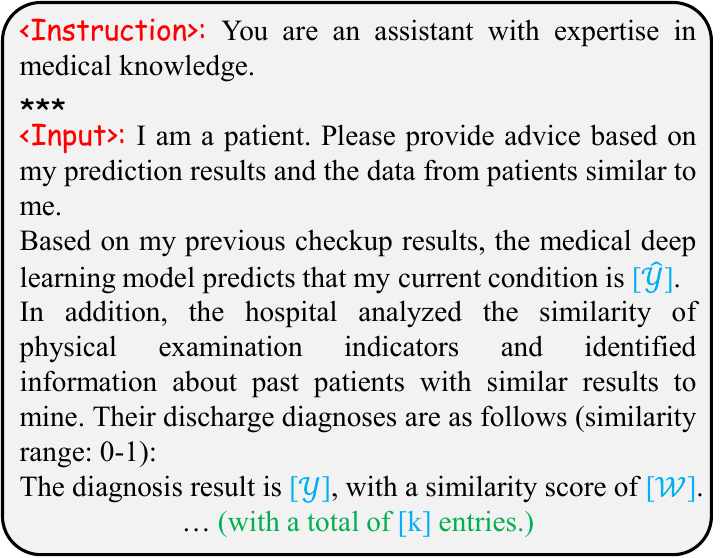}
    \caption{Example of the coarse-grained personalized prompt. The labels for similar patients are ground truth values, as these records come from past cases.}
    \label{fig:personalizedprompt}
\end{figure}

{\noindent\bf Coarse-grained Personalized Prompt.} Similar to previous works, our initial personalized prompt includes background and input data, as illustrated in Figure \ref{fig:personalizedprompt}. Specifically, the background part provides an overview of the input data and the downstream task, helping the LLM to understand the context and objectives. The input information is then populated with the personalized data extracted earlier, covering both the target patient's prediction outcomes and the recommendations from similar patients. Note that the obtained initial prompts already incorporate a coarse level of personalization since the input data is tailored to each individual, resulting in unique prompts for each patient.

{\noindent\bf Fine-grained Personalized Prompt.} To achieve a finer level of personalized prompts, we employ RL to refine the initial prompt. We design a word-level policy network that adjusts selected words to better guide LLMs in generating personalized responses. Since each initial prompt varies in length, {\modelname} operates at the word level, unlike previous works that treat the prompt as a single fixed input.

We formulate optimization as a Markov Decision Process \cite{mdp}. {\modelname} performs $n$ editing steps on the initial personalized prompt, where $n$ is a hyper-parameter controlling refinement depth. We define the state set as $\mathcal{S} = \{s_{0}, s_{1}, \cdots, s_{n-1}\}$, where $s_{i}, i \in [0, n-1]$ denotes the personalized prompt after the $i$-th optimization. Moreover, $s_0$ is the initial state, i.e., the initial prompt (Figure \ref{fig:personalizedprompt}). After that, {\modelname} introduces a policy network $f(\cdot)$ to determine the modification probability $\pi(e|s)$ for each word under a given state $s$. Formally, this can be described as follows: 
\begin{align}
    & \pi(e|s) = softmax(f(E)), \label{eq_prob1}\\
    & E = BERT(s) = (e_{1}, e_{2}, \cdots, e_{N_{s}}), \label{eq_bert}
\end{align}
where $softmax(\cdot)$ denotes normalization and $E$ is the embedding of the personalized prompt at state $s$. The personalized prompt is textual, thus, {\modelname} first encodes it using BERT \cite{bert}. Moreover, $e$ indicates word embeddings, where each embedding has the same dimension, in contrast, the number of words $N_{s}$ across states may vary. {\modelname} samples a word from the probability distribution $\pi(e|s)$ and removes it to obtain the next state. This editing process allows the policy network to explore different prompt variants and select those that better align the generated responses with physician guidance. This process is repeated $n$ times, each time deleting only one word, until reaching the final state $s_{n-1}$, which serves as the personalized prompt $P$.

Nonetheless, the transition probability relies heavily on word embeddings, resulting in limited personalization. In detail, while these embeddings include contextual information, they still lack sufficient granularity to differentiate between individual patients. To address this, {\modelname} incorporates patient information along with sentence representations. Formally, Eq.(\ref{eq_prob1}) can be reformulated as follows:
\begin{align}
    & \pi(e_i|s) = \frac{\exp\left(f(e_i \| \bar{E} \| \hat{x})\right)} {\sum_{j=1}^{N_s}\exp\left(f(e_j \| \bar{E} \| \hat{x})\right)}, \label{eq_prob2}\\
    & \bar{E} = mean(E) = \frac{e_{1}+e_{2}+\cdots+e_{N_s}}{N_s}, \label{eq_mean}
\end{align}
where $i \in [1,\dots,N_s]$ and $\|$ denotes concatenation. $\bar{E}$ captures the global semantics of the prompt, while $\hat{x}$ denotes the patient representation obtained from the encoder in Eq.(\ref{eq_encoder}). With this refinement, the prompt modification considers patient information and overall semantics, enabling the generation of a more personalized prompt. For simplicity, we still employ the MLP as the policy network.

After $n$ iterations, we obtain the personalized prompt $P$. We define the reward using BERTScore \cite{bertscore}, which is as follows:
\begin{align}
    & Reward = BS(\hat{r}, r) - BS(\hat{r}_{0}, r), \label{eq_reward}\\
    & \hat{r} = LLM_{r}(P), \label{eq_reply}
\end{align}
where $LLM_{r}$ indicates the LLM used to generate replies and $\hat{r}$ is its response to the personalized prompt, $\hat{r}_{0} = LLM_{r}(s_0)$ is the response to the initial personalized prompt. Moreover, $BS(\cdot)$ stands for BERTScore, and $r$ denotes the reference responses provided by physicians for each patient, which reflect personalized medical guidance in clinical practice. The reward is computed once based on the final refined prompt by comparing the generated response with the physician reference response using BERTScore. Since physician responses are personalized for each patient, this reward encourages the learned prompts to capture guidance for individual patients. The reward function also considers the downstream LLM, meaning that the personalized prompt is adaptively revised in alignment with $LLM_r$. As the personalized prompt is in text format, we avoid any modifications to $LLM_r$, enabling compatibility with proprietary LLMs. In this case, the loss function is defined as:
\begin{equation}\label{eq_loss}
    \mathcal{L} = - \sum_{i=0}^{n-1}log(\pi(e_{j} | s_{i})) \cdot Reward ,
\end{equation}
where $\pi(e_{j}|s_{i})$ is the probability of deleting word $e_j$ in the state $s_i$. Moreover, $j \in [1, N_{s_{i}}]$ denotes a chosen index, following the probability distribution $\pi(e|s_{i})$. Finally, we optimize our proposed {\modelname} by Adam \cite{adam}.

We further provide the pseudo-code of {\modelname}, training strategies, and an interactive web interface in Appendix \ref{modeldetails}.

\section{Experiments}

\subsection{Experimental Setup}

{\noindent\bf Datasets.} We collect real-world obstetrics data from multiple hospitals in Wuhan city and surrounding areas from 2020 to 2022. Details on data processing are provided in the Appendix \ref{datadetails}. The final dataset consists of 38,817 records for 2,373 pregnant patients, with each record containing 35 examination metrics. The patient with the most records has 45 entries, and each patient is assigned a label and receives medical recommendations, with a total of 12 categories. We partition the data by year: data from 2021 and earlier form the training set, the first half of 2022 as the validation set, and the remaining data as the test set. This dataset is suitable for our task since it contains longitudinal health records together with medical recommendations written by physicians for each patient, which are used both as the RL optimization signal and as reference responses in evaluation. We use prenatal examination records available before delivery and retrieve similar patients from the training set, avoiding future information leakage.

{\noindent \bf Baselines.} We evaluate {\modelname} on multiple proprietary LLMs and compare it with several fine-tuned lightweight LLMs to examine whether prompt learning on strong proprietary LLMs can outperform directly fine-tuning lightweight LLMs. Our goal is not to compare model capacity across different scales, but to evaluate prompt learning as a practical alternative when proprietary LLMs cannot be fine-tuned. We do not compare with prompt optimization methods, since they mainly optimize shared prompts for tasks rather than personalized prompts. The baselines are as follows:
\squishlist
    \item {\bf Proprietary LLMs.} We evaluate {\modelname} on several LLMs, including Gemini1.5-pro \cite{gemini1.5}, GLM4 \cite{chatglm}, GLM4-plus, GPT3.5-turbo \cite{gpt3}, and GPT4 \cite{gpt4}.
    \item {\bf Lightweight LLMs.} We compare several LLMs that are fine-tuned on the training data, including Llama3-8B \cite{llama3}, GLM4-9B \cite{chatglm}, Qwen2-7B \cite{qwen2}, and LLaVA1.5-7B \cite{llava1.5}.
\squishend

{\noindent\bf Experimental settings.} We evaluate {\modelname} on a server with an Intel Xeon Gold 5117 CPU, a Tesla V100 GPU (32 GB), and 256 GB RAM. The server uses Ubuntu 18.04 with CUDA 12.1, and our code is implemented in PyTorch 2.1.0. To reduce randomness, we report the average results of five runs.

For hyper-parameter settings, both the encoder and policy network in {\modelname} are MLPs, each with hidden layer dimensions of 256 and a dropout rate of 0.4. Specifically, the encoder has 2 layers with an output dimension of 128, while the policy network has 3 layers with an output dimension of 1. The number of similar patients $k$ and the number of prompt modification steps $n$ are both set to 10, with a learning rate of 0.005. For {\modelname}’s predictor and other fine-tuned LLMs, we follow a representative work \cite{llamafactory}.

\subsection{Performance Study}\label{performancestudy}

\begin{table*}[t]
\begin{center}
    \caption{Performance of {\modelname} based on proprietary LLMs.}
    \label{tab:llmperformance}
    \begin{tabular}{cc|cc|cc|ccc}
        \toprule
         \multicolumn{2}{c|}{\multirow{2}{*}{Model}} & \multirow{2}{*}{BLEU-4} & \multirow{2}{*}{ROUGE-L} & \multicolumn{2}{c|}{ROUGE-N} & \multicolumn{3}{c}{BERTScore} \\
        \cmidrule{5-9}
        & & & & 1 & 2 & Precision & Recall & F1 \\
        \midrule
        \multirow{2}{*}{Gemini1.5-pro} & before & 6.49 & 15.46 & 27.57 & 6.19 & 62.79 & 65.15 & 63.94\\
                                   & after & {\bf 10.65} & {\bf 22.39} & {\bf 37.01} & {\bf 13.67} & {\bf 69.54} & {\bf 70.75} & {\bf 70.13} \\
        \midrule
        \multirow{2}{*}{GLM4} & before & 8.39 & 19.46 & 30.93 & 9.95 & 63.06 & 64.95 & 63.99 \\
                                   & after & {\bf 9.42} & {\bf 21.31} & {\bf 34.67} & {\bf 11.57} & {\bf 66.80} & {\bf 69.51} & {\bf 68.12} \\
        \midrule
        \multirow{2}{*}{GLM4-plus} & before & 7.79 & 15.45 & 33.15 & 10.48 & 59.43 & 63.26 & 61.26 \\
                                   & after & {\bf 9.04} & {\bf 18.47} & {\bf 36.22} & {\bf 11.96} & {\bf 65.20} & {\bf 69.58} & {\bf 67.31} \\
        \midrule
        \multirow{2}{*}{GPT3.5-turbo} & before & 8.73 & 20.08 & 29.48 & 9.69 & 65.56 & 64.90 & 65.19 \\
                                      & after & {\bf 9.39} & {\bf 22.84} & {\bf 32.41} & {\bf 10.49} & {\bf 69.25} & {\bf 68.59} & {\bf 68.89} \\
        \midrule
        \multirow{2}{*}{GPT4} & before & 7.76 & 15.49 & 28.62 & 9.38 & 61.03 & 64.59 & 62.75 \\
                              & after & {\bf 10.88} & {\bf 22.69} & {\bf 39.03} & {\bf 14.30} & {\bf 69.80} & {\bf 71.21} & {\bf 70.50} \\
        \bottomrule
    \end{tabular}
\end{center}
\end{table*}

\begin{table*}[t]
\begin{center}
    \caption{Comparison with lightweight LLMs.}
    \label{tab:finetuneperformance}
    \begin{tabular}{c|cc|cc|ccc}
        \toprule
        \multirow{2}{*}{Model} & \multirow{2}{*}{BLEU-4} & \multirow{2}{*}{ROUGE-L} & \multicolumn{2}{c|}{ROUGE-N} & \multicolumn{3}{c}{BERTScore} \\
        \cmidrule{4-8}
        & & & 1 & 2 & Precision & Recall & F1 \\
        \midrule
        Llama3-8B  & 9.56 & {\bf 23.52} & 38.44 & 14.12 & 65.61 & 65.93 & 65.77 \\
        GLM4-9B & 8.79 & 19.72 & 36.40 & 12.02 & 64.37 & 64.29 & 64.33 \\
        Qwen2-7B & 8.03 & 19.78 & 36.64 & 12.18 & 64.38 & 64.22 & 64.30\\
        LLaVA1.5-7B & 7.45 & 16.61 & 33.26 & 13.59 & 61.30 & 63.29 & 62.28\\
        {\modelname} & {\bf 10.88} & 22.69 & {\bf 39.03} & {\bf 14.30} & {\bf 69.80} & {\bf 71.21} & {\bf 70.50} \\
        \bottomrule
    \end{tabular}
\end{center}
\end{table*}

\begin{figure}[t]
    \centering
    \includegraphics[width=0.99\linewidth]{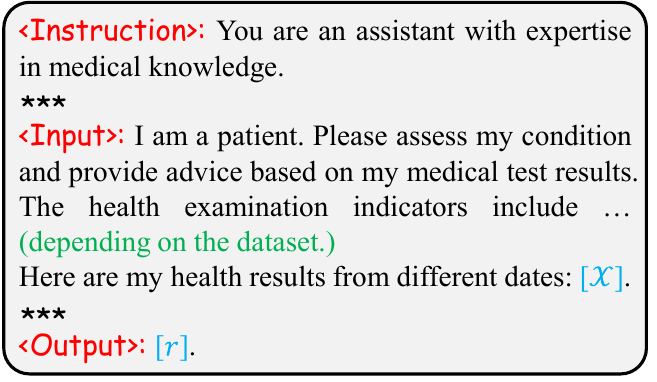}
    \caption{Example prompt for evaluation, where $<$Output$>$ is omitted for proprietary LLMs.}
    \label{fig:evaluateprompt}
\end{figure}

We evaluate whether {\modelname} improves alignment between generated responses and physician recommendations written for each patient, and compare with fine-tuned lightweight LLMs. To this end, we use the prompt shown in Figure \ref{fig:evaluateprompt}. For proprietary LLMs, we directly evaluate on the test data, as fine-tuning is not feasible. For lightweight LLMs, we fine-tune on the training data, followed by evaluation. We compare the outputs from each LLM with the physician reference responses, which reflect personalized guidance for each patient, using BLEU \cite{bleu}, ROUGE \cite{rouge}, and BERTScore \cite{bertscore}. These automatic metrics are mainly used to evaluate the lexical and semantic alignment between generated responses and physician reference responses. We further conduct human evaluation in Section \ref{humanevaluationstudy} to assess the quality of personalized healthcare guidance from an expert perspective.

From Table \ref{tab:llmperformance}, it is evident that {\modelname} consistently improves the performance of general LLMs. By leveraging personalized prompts, all evaluated LLMs exhibit performance improvements exceeding 10$\%$ on several metrics. In addition, the GPT4-based {\modelname} performs best, thus, we use GPT4 as the backbone in subsequent experiments.

To provide a more comprehensive evaluation, we also compare {\modelname} with fine-tuned LLMs, which are presented in Table \ref{tab:finetuneperformance}. We observe that {\modelname} achieves the best overall performance in this comparison. However, it falls short on ROUGE-L, which is expected since this metric relies heavily on word-level overlap. For the lightweight LLMs, their vocabulary is closer to the reference responses since they are trained directly on the physician replies in the training data. In this task, BERTScore is particularly important since it measures semantic similarity between generated responses and physician recommendations. Notably, {\modelname} achieves the highest BERTScore, indicating stronger semantic alignment with physician guidance. 

\begin{figure}[t]
    \centering
    \includegraphics[width=0.8\linewidth]{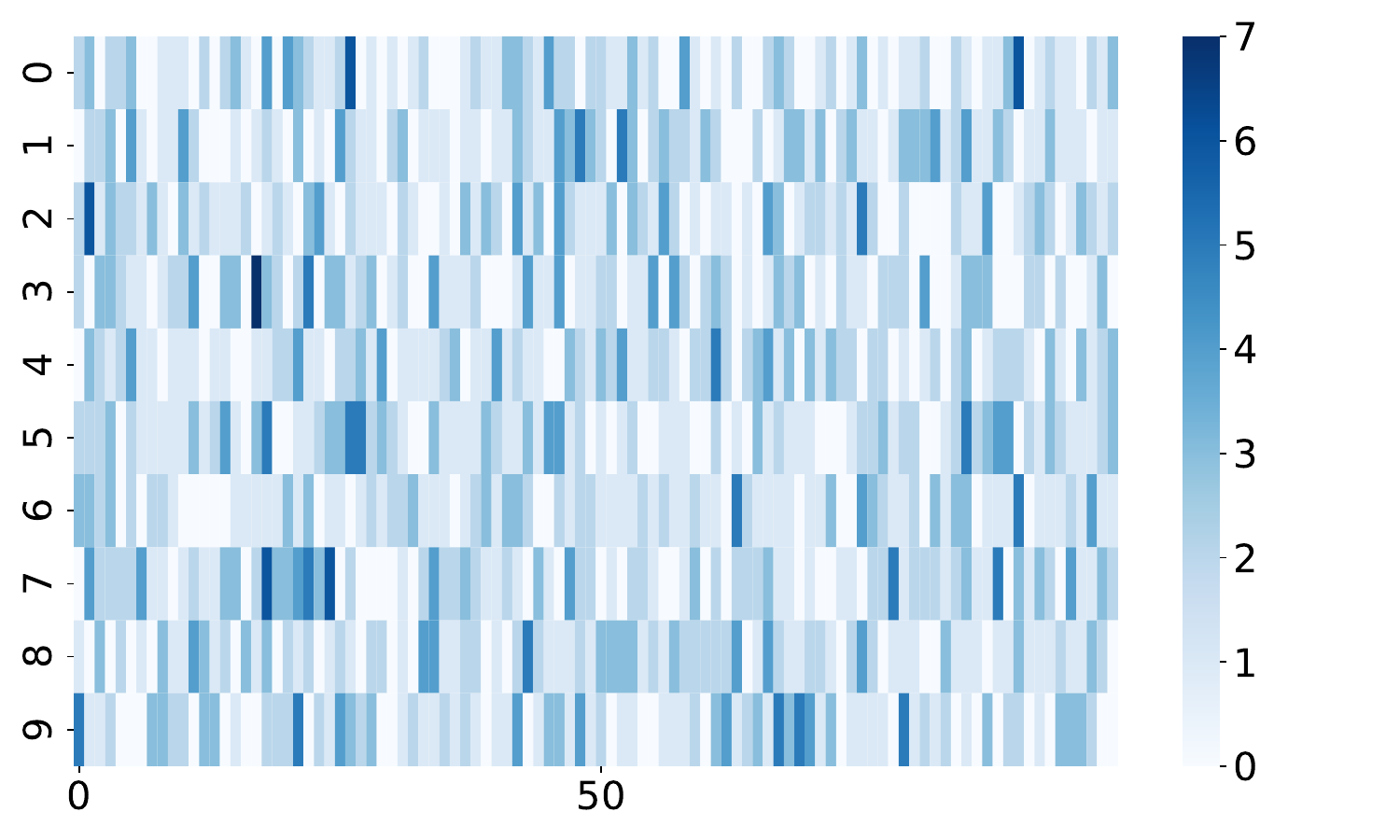}
    \caption{Heatmap of word modification counts, where the X-axis represents prompt word indices and the Y-axis represents refinement iterations.}
    \label{fig:personalization}
\end{figure}

\begin{table*}[t]
\centering
\caption{Examples for the personalization study of {\modelname}.}
\label{tab:case_study}
\resizebox{0.99\textwidth}{!}{
\begin{threeparttable}
\begin{tabular}{c|l}
\toprule
Case & Details \\
\midrule
\multirow{2}{*}{Case 1} & \textbf{Similar cases:} Uterine scarring and gestational diabetes.\\
& \textbf{Response:} ``Your examination results are similar to multiple prior cases, especially those involving uterine scarring and gestational diabetes...''\\
\midrule
\multirow{2}{*}{Case 2} & \textbf{Similar cases:} Gestational hypertension.\\
& \textbf{Response:} ``Although the prediction result suggests premature rupture of membranes, gestational hypertension may also need attention...''\\
\midrule
\multirow{2}{*}{Case 3} & \textbf{Similar cases:} Premature rupture of membranes.\\
& \textbf{Response:} ``Among prior cases with examination results similar to yours, premature rupture of membranes appears frequently...''\\
\bottomrule
\end{tabular}
\begin{tablenotes}
\item All cases share the same predicted condition (PROM). ``Similar cases'' summarize the dominant diagnosis categories among retrieved similar patients. Response segments are directly extracted from {\modelname}'s outputs, with irrelevant parts omitted.
\end{tablenotes}
\end{threeparttable}
}
\end{table*}

\begin{table*}[t]
\begin{center}
  \caption{Ablation study of {\modelname}.}
  \label{tab:ablation}
  \begin{threeparttable}
  \begin{tabular}{c|ccc|cc|cc|ccc} 
    \toprule
     \multicolumn{4}{c|}{Model} & \multirow{2}{*}{BLEU-4} & \multirow{2}{*}{ROUGE-L} & \multicolumn{2}{c|}{ROUGE-N} & \multicolumn{3}{c}{BERTScore} \\
     \cmidrule{1-4} \cmidrule{7-11}
     ID & SP & PP & PR & & & 1 & 2 & Precision & Recall & F1 \\
     \midrule
     1 & \ding{55} & \ding{51} & \ding{51} & 8.42 & 18.39 & 34.01 & 10.85 & 61.10 & 65.01 & 62.99 \\
     2 & \ding{51} & \ding{55} & \ding{51} & 9.29 & 21.38 & 34.54 & 11.51 & 65.92 & 68.61 & 67.24 \\
     3 & \ding{51} & \ding{51} & \ding{55} & 7.91 & 17.68 & 33.07 & 10.39 & 63.62 & 66.09 & 64.83 \\
     {\modelname} & \ding{51} & \ding{51} & \ding{51} & {\bf 10.88} & {\bf 22.69} & {\bf 39.03} & {\bf 14.30} & {\bf 69.80} & {\bf 71.21} & {\bf 70.50}\\
     \bottomrule
  \end{tabular}
  \begin{tablenotes}
        \item ID indicates different ablation variants, and '\ding{51}' and '\ding{55}' indicate whether a module is included.
  \end{tablenotes}
  \end{threeparttable}
\end{center}
\end{table*}

\subsection{Personalization Study}

To illustrate how {\modelname} generates personalized prompts for different patients, we analyze the word modification behavior of the policy network on the test set. Specifically, we record the indices of words removed during each refinement iteration. Due to space constraints, we report the modification counts for the first 100 words.

As shown in Figure \ref{fig:personalization}, the frequency of word modifications varies significantly across different indices. This indicates that the policy network adapts the refinement process according to the prompt context and patient information, producing different prompts for different patients.

To further examine whether such prompt differences lead to personalized healthcare guidance, we conduct a targeted case study using three patients with the same predicted condition, i.e., {\em premature rupture of membranes} (PROM). Table \ref{tab:case_study} presents response segments generated by {\modelname} that reflect the influence of peer-informed information. Although their predicted labels are identical, their examination histories lead to different retrieved similar patients, providing different peer-informed information for generating healthcare guidance. 

These cases show that {\modelname} does not simply generate guidance based on the predicted condition. Instead, it incorporates patient examination histories and peer-informed information, resulting in different healthcare guidance even for patients with the same predicted condition.

\subsection{Ablation Study}\label{ablationstudy}

We further assess the contributions of each module in {\modelname}. In detail, we examine the roles of {\em self-informed personalization} (SP), {\em peer-informed personalization} (PP), and {\em prompt refinement} (PR) in guiding LLMs to generate personalized responses.

Table \ref{tab:ablation} shows that all modules of {\modelname} contribute to its performance, confirming the effectiveness of the overall design. Compared with the other variants, the impact of PP (Variant 2) is smaller, since PP relies on the similarity of medical records, which may not always reflect true clinical similarity. As a result, even without this component, the model can still achieve reasonable performance.

\begin{table}[t]
\begin{center}
    \caption{Human evaluation study of {\modelname}.}
    \label{tab:humanevaluation}
    \begin{tabular}{c|ccc}
        \toprule
        Model & GLM4 & GPT3.5-turbo & {\modelname} \\
        \midrule
        Vote Share & 5.56 & 8.32 & {\bf 86.12} \\
        Wins & 1 & 2 & {\bf 97} \\
        \bottomrule
    \end{tabular}
\end{center}
\end{table}

\subsection{Human Evaluation Study}\label{humanevaluationstudy}

To validate the practical utility of {\modelname}, we conduct a human evaluation study. Given the high cost of manual assessment, we randomly sample 100 cases from our dataset and select GLM4 and GPT3.5-turbo as representative proprietary LLM baselines. The evaluation was conducted by 25 evaluators with medical backgrounds, including 7 licensed obstetricians, 3 PhD students, and 15 medical master students. For each case, participants are presented with anonymized outputs from different LLMs in randomized order and asked to identify the most appropriate healthcare guidance according to clinical relevance, usefulness, and clarity. The final preference is determined by majority voting among the participants.

As shown in Table \ref{tab:humanevaluation}, {\modelname} is preferred in 97 out of 100 cases, outperforming the baselines and demonstrating the effectiveness of personalized healthcare guidance. Additional case studies are provided in Appendix \ref{casestudy}.

\begin{figure}[t]
    \centering
    \subfigure[Analysis of $k$.]{\includegraphics[width = 0.493\linewidth]{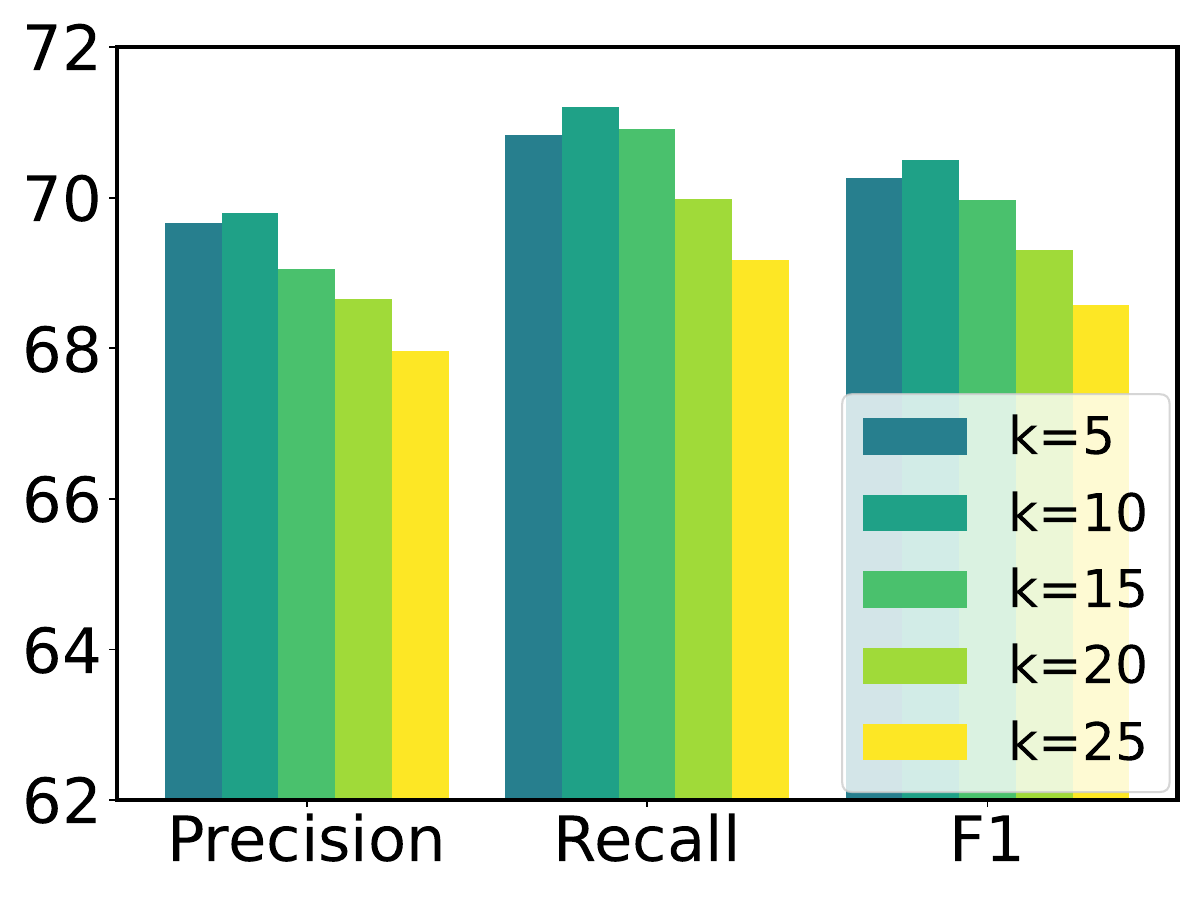}}
    \subfigure[Analysis of $n$.]{\includegraphics[width = 0.493\linewidth]{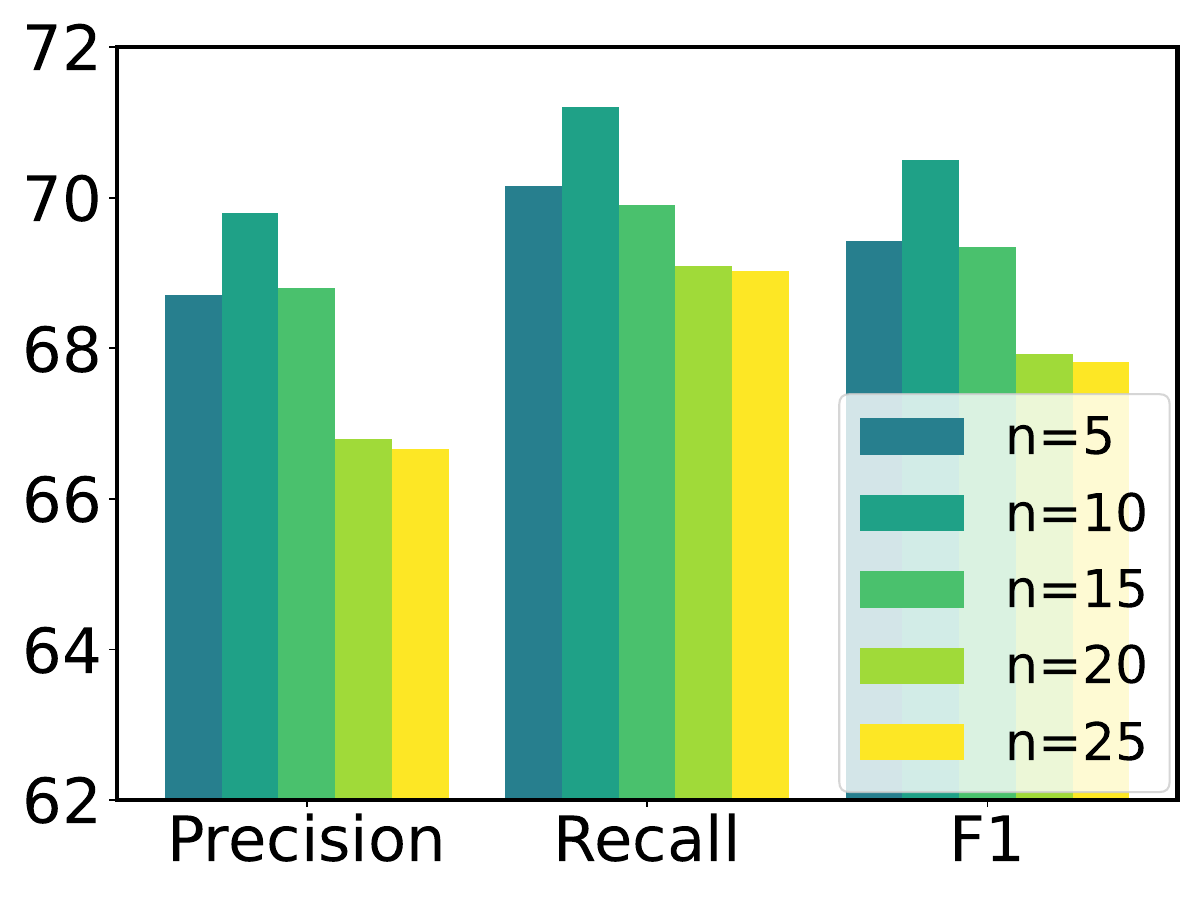}}
    \caption{Hyper-parameter study. The three groups correspond to BERTScore precision, recall, and F1, and each group contains results for different $k$ or $n$.}
    \label{fig:hyperparameter}
\end{figure}

\subsection{Hyper-parameter Study}\label{hyperparameterstudy}

We also investigate the impact of hyper-parameters on performance, focusing on the number of similar patients $k$ and prompt modification iterations $n$.

The results in Figure \ref{fig:hyperparameter} suggest that as $k$ or $n$ increases, the performance of {\modelname} first improves and then declines. Increasing $k$ beyond a certain point may introduce patients with lower similarity, whose records may introduce irrelevant or conflicting information, thereby weakening personalization. Likewise, for $n$, excessive modifications to the prompt may distort its semantics and reduce the reliability of the guidance. Moreover, $n$ has a more significant impact on performance, which is consistent with the importance of prompt refinement observed in the ablation study. Based on these observations, we set $k$ and $n$ to 10 in our experiments.

\section{Conclusion}

We study personalization for LLMs in healthcare and propose the {\modelname} framework. {\modelname} constructs prompts tailored to each patient from their health data and refines these prompts through RL to better align with physician recommendations. The resulting personalized prompts guide LLMs to generate customized responses. Moreover, {\modelname} can be applied to various LLMs, including proprietary models, without directly sending raw patient records. Extensive experiments, evaluated using both automatic metrics and human experts, demonstrate the effectiveness of {\modelname} and highlight its potential for advancing adaptive healthcare LLMs.


\section*{Limitations}

While {\modelname} explores personalized prompt learning for healthcare LLMs and demonstrates promising results, several limitations remain.

\squishlist
    \item {\bf Dataset.} {\modelname} learns personalized prompts by aligning generated responses with physician recommendations, while public healthcare datasets rarely contain both longitudinal patient records and physician guidance. We thus evaluate {\modelname} on a single real-world obstetrics and gynecology dataset. Although this dataset reflects realistic clinical practice, it may not cover the full diversity of medical scenarios. Evaluating {\modelname} on broader datasets from different hospitals could better assess its generalizability.
    \item {\bf Prompt editing.} {\modelname} refines prompts using a deletion-based editing strategy, which keeps the editing space stable and allows RL to explore prompt modifications efficiently. However, richer operations such as insertion or replacement could provide more flexible prompt optimization. In addition, deletion may occasionally produce prompts that appear semantically incoherent. Prior studies \cite{incoherent,rlprompt} report that LLMs can still generate effective responses under such prompts, possibly since LLMs interpret prompt signals differently from human readers.
\squishend

\section*{Ethical Consideration}

We conduct this study in collaboration with Tongji Medical College of HUST and Maternal and Child Health Hospital of Hubei Province, using fully anonymized retrospective medical records, with all identifiable patient and physician information removed prior to analysis. The dataset contains no direct personal identifiers, and all data are handled in accordance with standard privacy protection practices. Moreover, {\modelname} avoids sending raw records to proprietary LLMs, and all data are used solely for academic research aimed at improving healthcare applications.

\section*{Acknowledgements}
The work is supported by the National Natural Science Foundation of China~(No.62127808).

\clearpage


\bibliography{reference}
\clearpage

\appendix

\section{Appendix}
\setcounter{table}{0}
\setcounter{figure}{0}
\renewcommand{\thetable}{A\arabic{table}}
\renewcommand{\thefigure}{A\arabic{figure}}

\begin{figure}[!ht]
    \centering
    \includegraphics[width=0.95\linewidth]{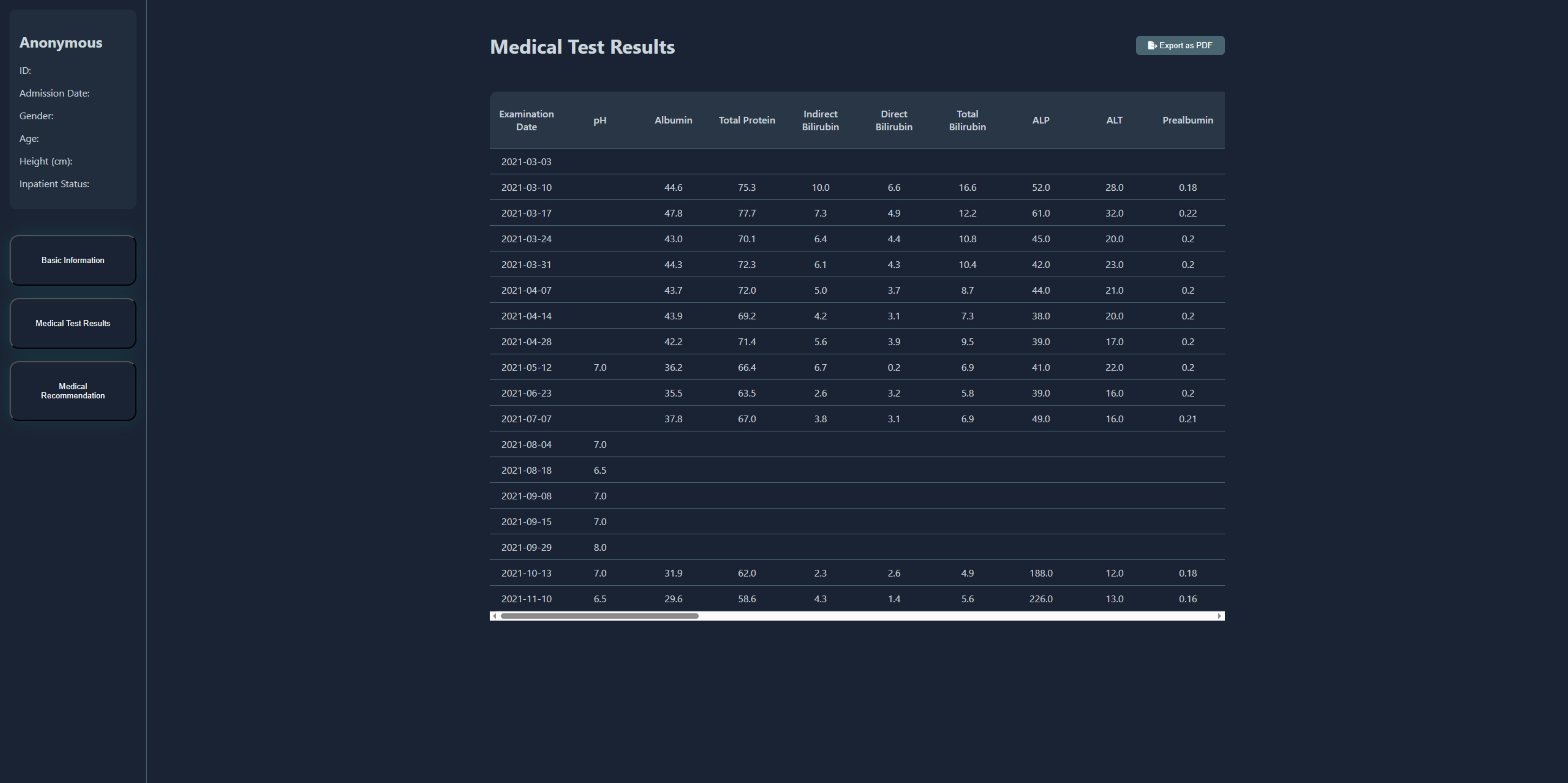}
    \caption{Web interface of {\modelname}. This page shows a patient’s medical test results.}
    \label{fig:webtestresult}
\end{figure}

\subsection{Details of {\modelname}}\label{modeldetails}

To further illustrate {\modelname}, we provide its pseudo-code in Algorithm \ref{modelalgorithm}. The predictor is pre-trained to ensure effective personalized prompt construction. For test data, similar patients are selected from the training set to build personalized prompts, which reflects real clinical settings where hospitals maintain historical records.

We also present an interactive Web interface to demonstrate the practical use of {\modelname}. Figure \ref{fig:webtestresult} shows a representative page displaying a patient’s historical medical test results across different dates. The top-left corner presents the basic information of the patient, including gender and height. The left panel allows users to switch between viewing basic information, medical test results, and recommendations generated by {\modelname}. The interface has been briefly tested by internal users, suggesting that {\modelname} can guide the LLM to produce tailored responses for individual patients.

\begin{algorithm}[!ht]
\caption{The Proposed \modelname{}}\label{modelalgorithm}
\KwIn{Training set $\mathcal{H}^{tr} = \{\mathcal{X}^{tr}, \mathcal{Y}^{tr}, \mathcal{R}^{tr}\}$; Test set $\mathcal{H}^{te} = \{\mathcal{X}^{te}\}$; Number of epochs $E$; Number of similar patients $k$; Number of states $n$;}
\SetKw{Parameter:}{}
\KwOut{Responses of test set $\hat{R}$}
$\backslash\ast$ Pretraining $\ast\backslash$ \\
Fine-tune $LLM_{p}$ using $\mathcal{X}^{tr}$ and $\mathcal{Y}^{tr}$; \\
$\backslash\ast$ Training $\ast\backslash$ \\
$\hat{\mathcal{Y}}^{tr} = LLM_{p}(\mathcal{X}^{tr})$; Eq.(1) \\
\For{$epoch = 1 \cdots E$}{
    \For{$x \in \mathcal{X}^{tr}$}{
        $x' = padding(\sum^{N_{x}}_{j=1} \|x_{j})$; Eq.(3) \\
        $\hat{x} = encoder(x')$; Eq.(2) \\
        \scalebox{0.9}{Obtain $k$ most similar patients}; Eq.(4) \\
        \scalebox{0.9}{Construct initial personalized prompt}; \\
        \scalebox{0.9}{Initialize the state $s_{0}$}; \\
        \For{$i = 0 \cdots n-2$}{
            $E = BERT(s_{i})$; Eq.(6) \\
            $\bar{E} = \frac{e_{1} + e_{2} + \cdots + e_{N_{s}}}{N_{s}}$; Eq.(8)\\
            \scalebox{0.95}{$\pi(e_j|s_i)=\frac{\exp\left(f(e_j\|\bar{E}\|\hat{x})\right)}{\sum_{k=1}^{N_{s_i}}\exp\left(f(e_k\|\bar{E}\|\hat{x})\right)}$; Eq.(7)} \\
            \scalebox{0.95}{$s_{i+1} \leftarrow$ Modify $s_i$ based on $\pi(e|s_i)$}; \\
        }
        $P = s_{n-1}$; \\
        $\hat{r} = LLM_{r}(P)$; Eq.(10) \\
    }
    \scalebox{0.9}{$Reward = BS(\hat{\mathcal{R}}, \mathcal{R}^{tr}) - BS(\hat{\mathcal{R}}_{0}, \mathcal{R}^{tr})$; Eq.(9)} \\
    \scalebox{0.94}{$\mathcal{L} = - \sum_{i=0}^{n-1}log(\pi(e_{j}|s_{i})) \cdot Reward$; Eq.(11)} \\
    Update through back propagation based on the loss $\mathcal{L}$; \\
}
$\backslash\ast$ Inferring $\ast\backslash$ \\
$\hat{\mathcal{Y}}^{te} = LLM_{p}(\mathcal{X}^{te})$; Eq.(1) \\
\For{$x \in \mathcal{X}^{te}$}{
    \scalebox{0.99}{Obtain $k$ similar patients from the training set}; \\
    Construct initial personalized prompt; \\
    $P \leftarrow$ Modify initial prompt $n$ times; \\
    $\hat{r} = LLM_{r}(P)$; Eq.(10) \\
}
{\bf return} Responses of test set $\hat{R}$
\end{algorithm}

\subsection{Dataset Details}\label{datadetails}

We collect and process obstetrics and gynecology data from multiple hospitals in Wuhan city and surrounding areas, covering records from 2020 to 2022. The dataset includes longitudinal medical examination records of pregnant women from admission to discharge. It contains 194,345 entries, each with 302 examination indicators. Discharge diagnoses are used as labels, while physician recommendations serve as reference responses. {\em The use of this dataset complies with ethical standards, as described in the Ethical Consideration.}

{\bf\noindent Examination indicators.} We analyze the dataset and select 35 measurable examination indicators due to data sparsity and the textual nature of some indicators. The selected indicators include pH level, albumin, total protein, indirect bilirubin, direct bilirubin, total bilirubin, alkaline phosphatase, alanine aminotransferase, prealbumin, total bile acid, large platelet count, plateletcrit, large platelet ratio, mean platelet volume, platelet distribution width, red blood cell distribution width (coefficient of variation and standard deviation), basophil count, eosinophil count, monocyte count, lymphocyte count, neutrophil count, basophil ratio, eosinophil ratio, monocyte ratio, lymphocyte ratio, neutrophil ratio, mean corpuscular hemoglobin concentration, mean corpuscular hemoglobin, hematocrit, white blood cell count, platelet count, hemoglobin, mean corpuscular volume, and red blood cell count.

{\bf\noindent Labels.} The original dataset contains 455 conditions plus a "normal" category. We filter out conditions with insufficient samples and retain 12 labels: normal, uterine scar, acute fetal distress (heart type), pregnancy complicated with placental dysfunction, gestational hypertension, premature rupture of membranes, pregnancy complicated with hypothyroidism, uterine rupture, acute fetal distress (amniotic fluid type), umbilical cord entanglement, thalassemia, and gestational diabetes.

{\bf\noindent Final dataset.} After removing incomplete records and retaining the selected indicators, the final dataset contains 38,817 entries from 2,373 pregnant women. The dataset includes 56 patients from 2020, 1,638 from 2021, and 679 from 2022. We split the dataset by year: data from 2021 and earlier form the training set, the first half of 2022 as the validation set, and the remaining data as test set.

\begin{table}[t]
\begin{center}
    \caption{Predictor study.}
    \label{tab:predictor}
    \begin{tabular}{c|cc}
        \toprule
        Model & Micro-F1 & Macro-F1\\
        \midrule
        LSTM  &  21.60 & 5.51 \\
        CNN  &  20.89 & 6.42 \\
        ROCKET  & 36.05 & 20.58 \\
        Transformer & 41.12 & 24.58 \\
        Llama3-8B  & 79.41 & 62.94 \\
        GLM4-9B & 65.63 & 56.12 \\
        Qwen2-7B & 80.05 & 69.86\\
        LLaVA1.5-7B & {\bf 82.58} & {\bf 74.13} \\
        \bottomrule
    \end{tabular}
\end{center}
\end{table}

\subsection{Predictor Study}\label{predictorstudy}

We further study the choice of predictor. Patient data can essentially be viewed as time series data, so we evaluate both LLMs and representative time series methods, including CNN \cite{cnn}, LSTM \cite{lstm}, ROCKET \cite{rocket}, and Transformer \cite{transformer}. Since the number of medical visits varies across patients, zero padding is applied to align data. 

Table \ref{tab:predictor} shows that fine-tuned LLMs consistently outperform traditional time series methods by a large margin. This may be since healthcare data are sensitive to sequence padding, which can affect the performance of time series methods. Among the LLMs, LLaVA1.5-7B achieves the best performance. Although it is a multimodal model, using only its text input already yields strong results. Thus, we adopt it as the predictor in our work.

\subsection{Case Study}\label{casestudy}

\begin{figure*}[t]
    \centering
    \includegraphics[width=0.86\linewidth]{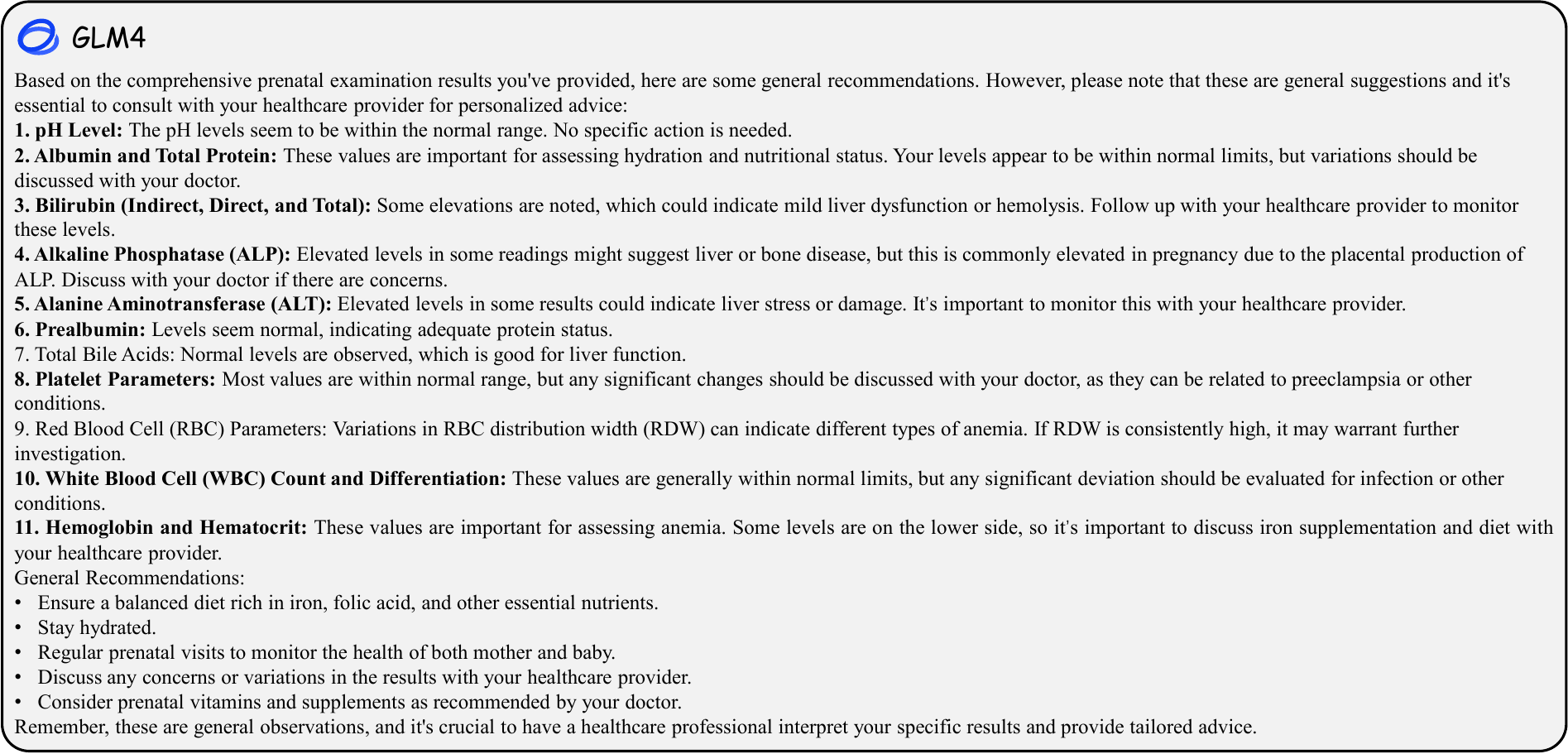}
    \caption{GLM4 response in the GDM case study.}
    \label{fig:glmgdm}
\end{figure*}

\begin{figure*}[t]
    \centering
    \includegraphics[width=0.86\linewidth]{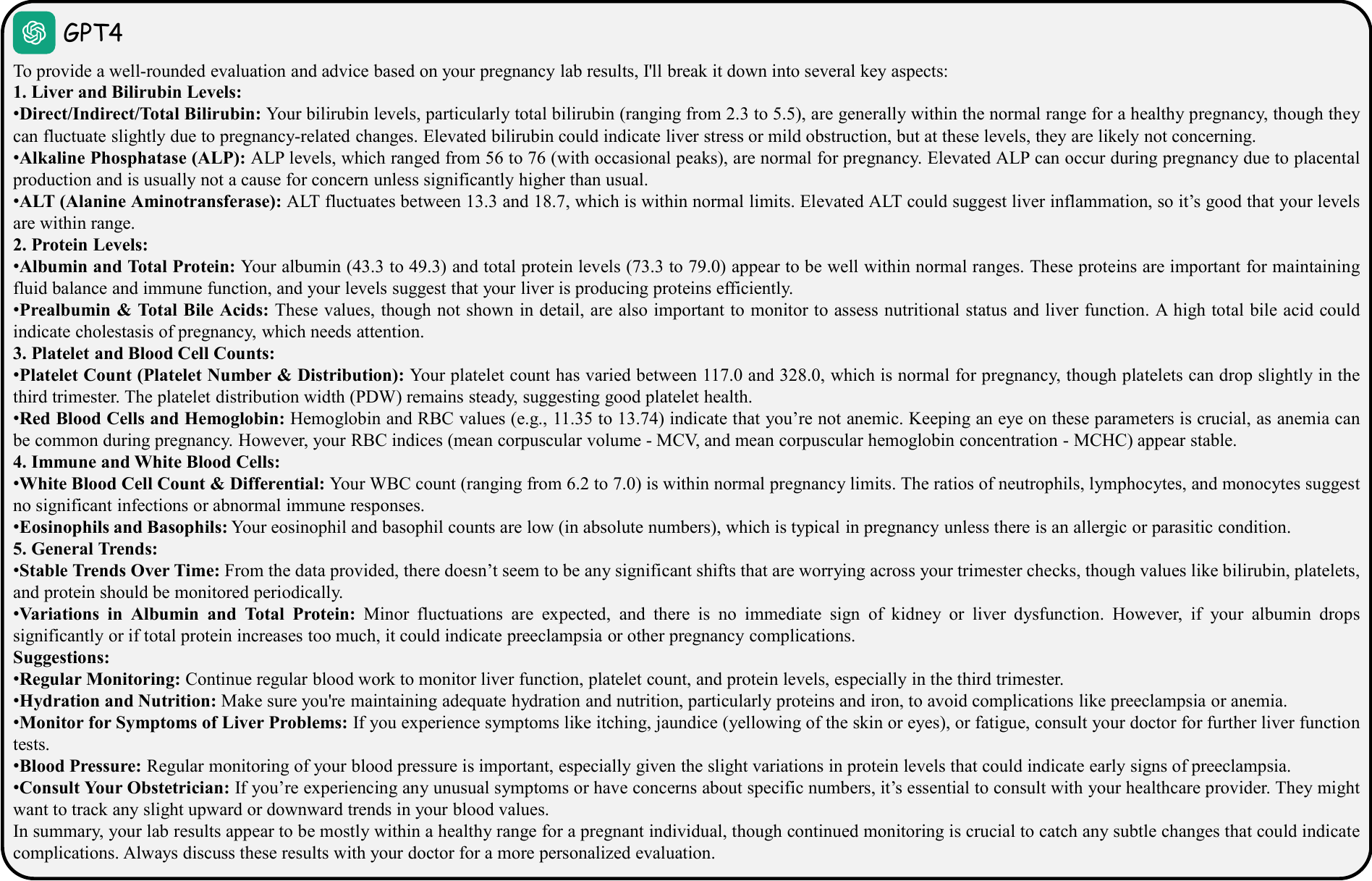}
    \caption{GPT4 response in the GDM case study.}
    \label{fig:gptgdm}
\end{figure*}

\begin{figure*}[t]
    \centering
    \includegraphics[width=0.86\linewidth]{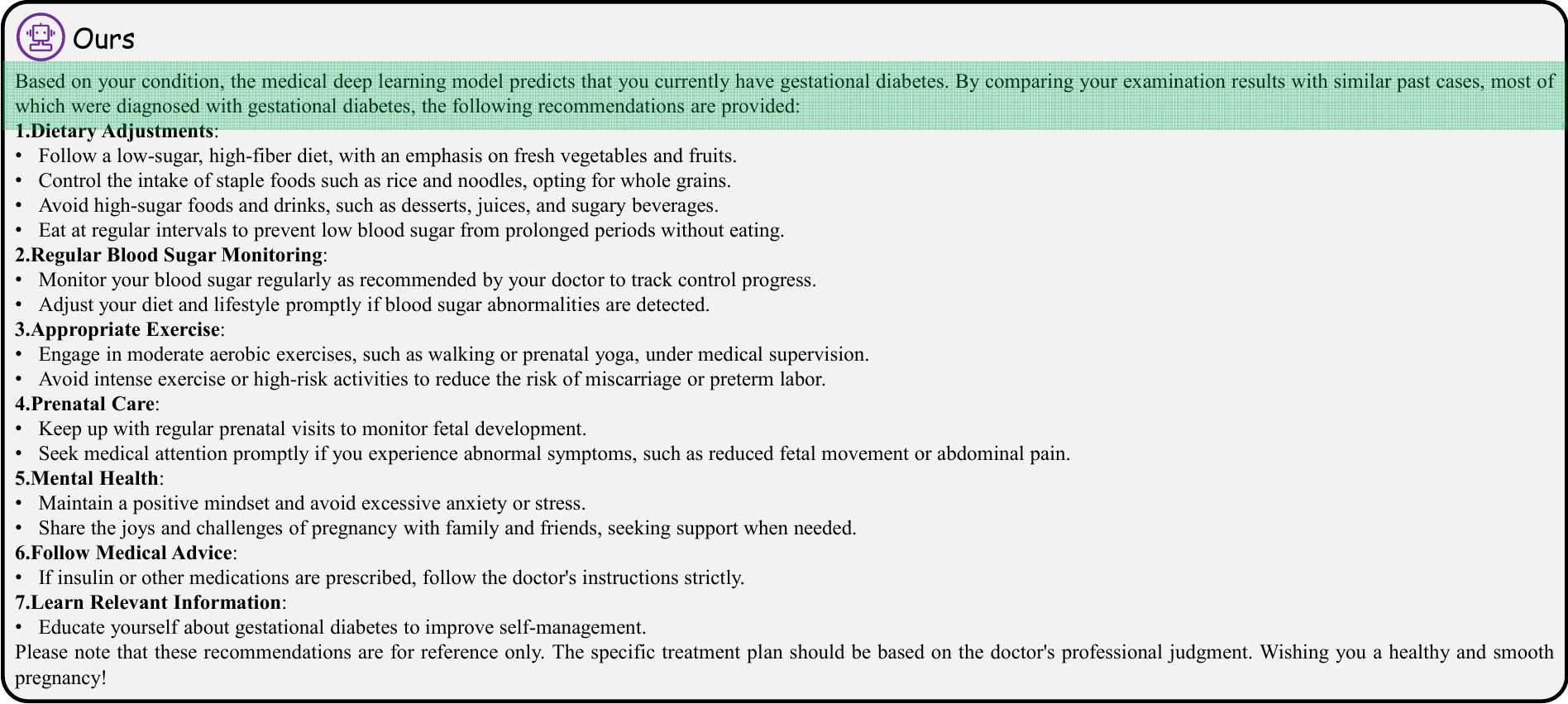}
    \caption{{\modelname} response in the GDM case study.}
    \label{fig:oursgdm}
\end{figure*}

\begin{figure*}[t]
    \centering
    \includegraphics[width=0.86\linewidth]{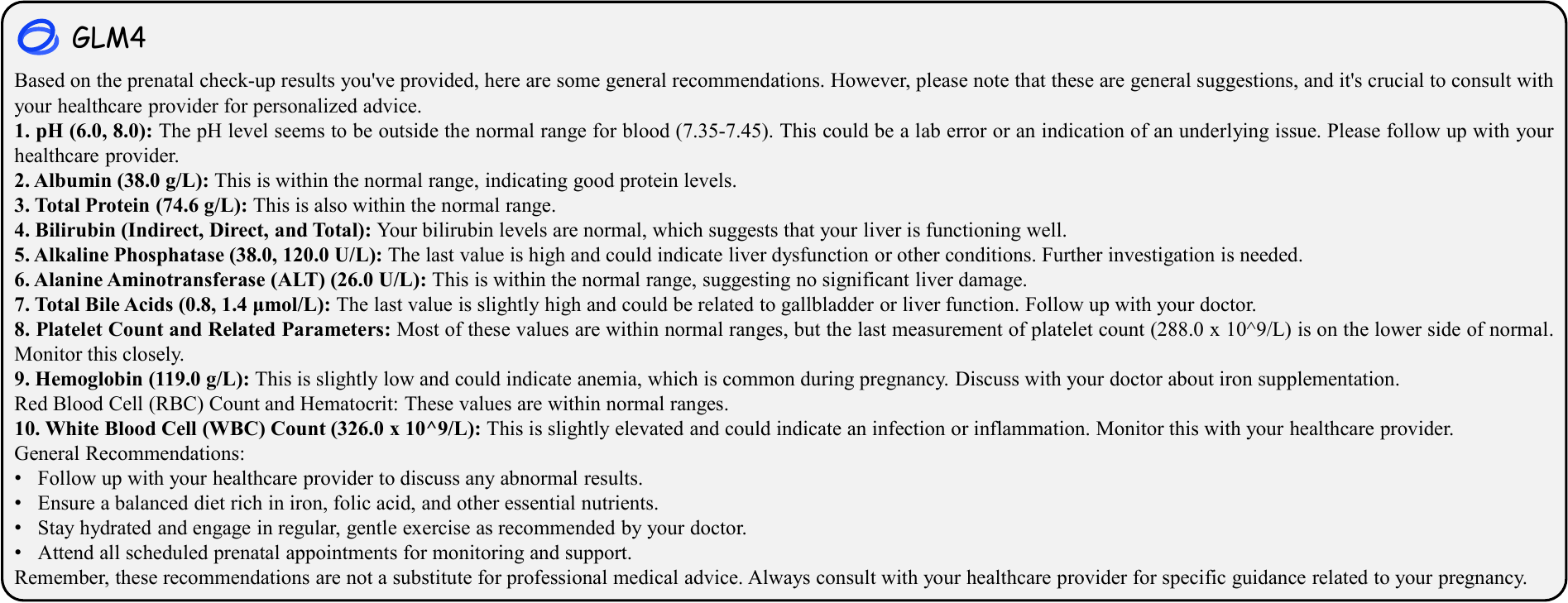}
    \caption{GLM4 response in the normal case study.}
    \label{fig:glmnormal}
\end{figure*}

\begin{figure*}[t]
    \centering
    \includegraphics[width=0.86\linewidth]{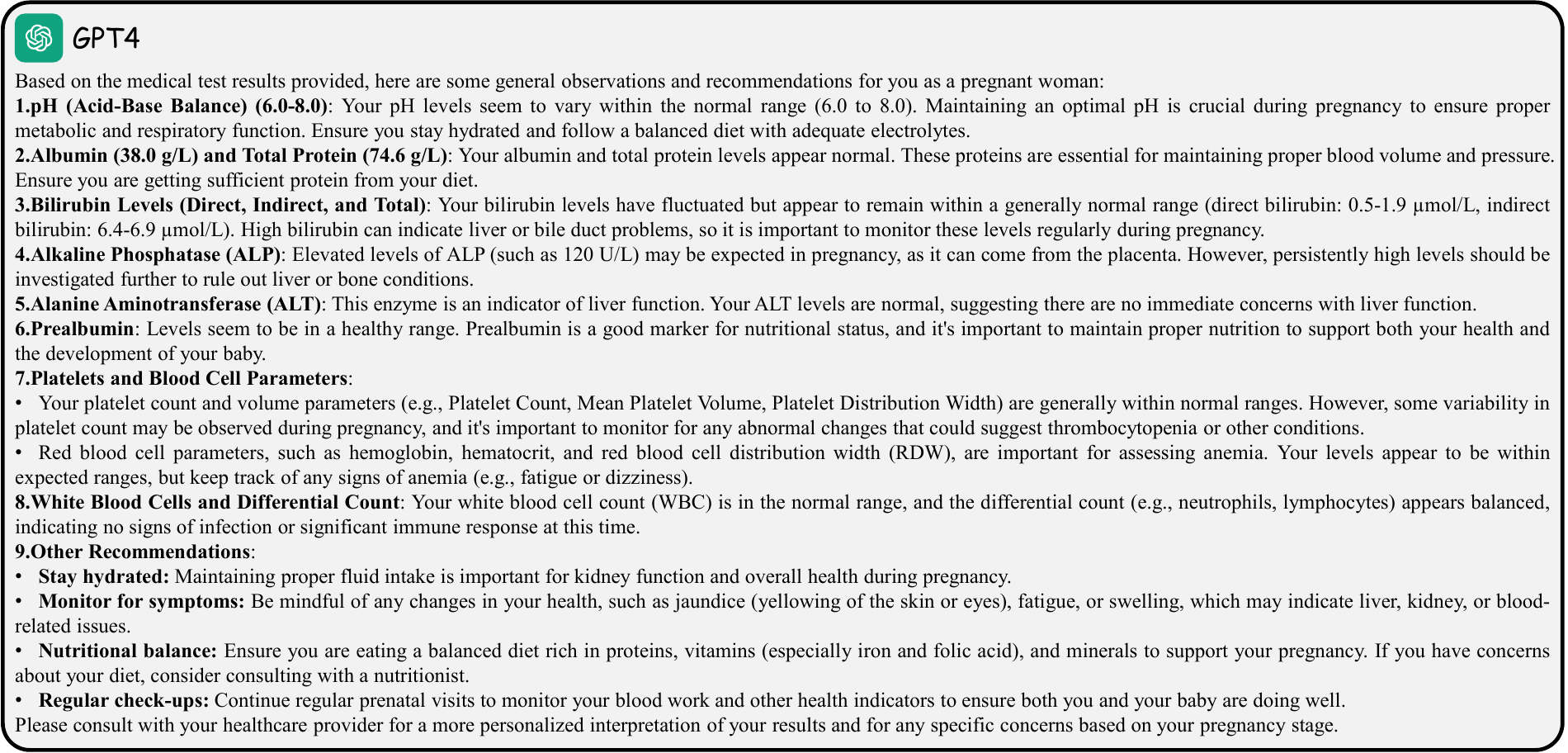}
    \caption{GPT4 response in the normal case study.}
    \label{fig:gptnormal}
\end{figure*}

\begin{figure*}[t]
    \centering
    \includegraphics[width=0.86\linewidth]{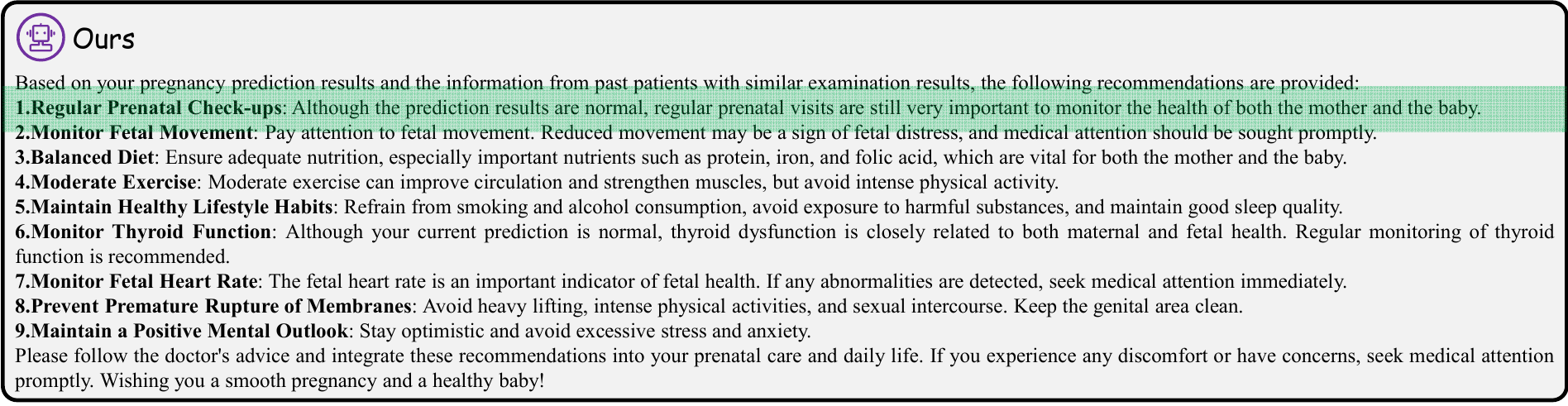}
    \caption{{\modelname} response in the normal case study.}
    \label{fig:oursnormal}
\end{figure*}

Here we provide several cases to illustrate how the proposed {\modelname} guides LLMs to generate personalized responses. We compare the responses of GLM4 \cite{chatglm} and GPT4 \cite{gpt4}. Specifically, we present cases for two different patient types, a pregnant woman diagnosed with {\em Gestational Diabetes Mellitus} (GDM) (Figures \ref{fig:glmgdm}, \ref{fig:gptgdm}, and \ref{fig:oursgdm}) and a pregnant woman with normal conditions (Figures \ref{fig:glmnormal}, \ref{fig:gptnormal}, and \ref{fig:oursnormal}).

The results show that {\modelname} can accurately identify the condition of patients (highlighted in green) and provide more tailored recommendations, demonstrating its ability to capture individual patient characteristics. In contrast, GLM4 and GPT4 tend to rely on individual numerical metrics and may fail to interpret the overall health status of patients. As a result, their responses are often more generalized. For example, in the case of the normal patient, both GLM4 and GPT4 incorrectly suggest potential conditions, which further highlights their limited ability to provide personalized healthcare guidance. These observations validate the effectiveness of {\modelname} in generating patient-specific recommendations and suggest its potential for developing more personalized LLM-based healthcare systems.

\end{document}